\documentclass{article}

\usepackage[T1]{fontenc}
\usepackage{newtxtext,newtxmath}

\usepackage{spconf,amsmath,graphicx,hyperref}
\usepackage{xcolor}
\usepackage{threeparttable}
\usepackage{fontawesome5}

\def\red{\textcolor{black}}

\title{Tuning-free Instruction-based Video Editing Via Structural Noise Initialization and Guidance}
\author{Song Wu}
\name{$\text{Song Wu}^{1~(*)}$, $\text{Xinyu Chen}^{1,2,3~(*)}$,  $\text{Qian Wang}^{1~(\text{\faEnvelope})}$, $\text{Liang Li}^{1}$, $\text{Zili Yi}^{2,3}$, and $\text{Junlan Feng}^{1~(\text{\faEnvelope})}$\thanks{*Equal contribution. This work was completed by Xinyu Chen during his internship at JIUTIAN Research. \faEnvelope~Corresponding authors.}}
\address{${}^1$JIUTIAN Research, China Mobile \ ${}^2$School of Intelligence Science and Technology, Nanjing University \\ ${}^3$State Key Laboratory of Novel Software Technology, Nanjing University \\ \{wusong, wangqian, fengjunlan\}@cmjt.chinamobile.com }
\begin{document}
%
\maketitle
\begin{abstract}
Video editing poses a significant challenge. While a series of tuning-free methods circumvent the need for extensive data collection and model training, they often underutilize the rich information embedded within noisy latent, leading to unsatisfactory results. To address this, we propose a \textit{tuning-free, instruction-based} video editing framework. We approach video editing from the perspective of noisy latent: we design a Structural Noise Initialization Strategy (SNIS) to secure a superior editing starting point by assigning higher noise levels to edited regions (to facilitate content change) and lower noise levels to unedited regions (to maintain content consistency). We introduce a Noise Guidance Mechanism (NGM), which leverages the video prior in the generative model and effectively integrates rich information within the noisy latent to guide the denoising process, thereby preserving unedited content and overall visual coherence. Experiments show that our proposed method achieves better visual quality and state-of-the-art performance.

\end{abstract}
\section{Introduction} 
\vspace{-0.1cm}
\label{sec:intro}
Video editing is a vital task in computer vision with implications for industries ranging from filmmaking to social networks. Its goal is to achieve harmonious coordination between the edited and unedited areas and retain unedited content while following the user instructions to complete the editing. Due to the lack of high-quality video editing pairs and instruction datasets, researchers focus on expanding video editing capabilities based on visual generation models. 

Early works customize image editing techniques~\cite{InstructPix2Pix, P2P, Plug-and-Play} for video editing tasks, such as introducing spatio-temporal attention mechanisms~\cite{Text2Video-Zero, FateZero, Pix2Video, Tune-A-Video} or additional motion information~\cite{TokenFlow, VideoShop, DIVE, VideoSwap} into the image generation model. Although tailored techniques alleviate the lack of temporal information in the image generation model, the edited videos still face temporal inconsistency. To transform the generation capabilities of video generation models into editing capabilities, subsequent training-based works~\cite{VACE, MiniMax} require careful dataset curation to fine-tune the video generation model~\cite{HunyuanVideo, CogVideoX, WanX}. In contrast, tuning-free methods~\cite{AnyV2V, V2Edit} attempt to tailor inversion mechanisms and attention manipulation techniques. Due to different corruption rates of high- and low-frequency information during diffusion~\cite{FreeInit}, edited and unedited regions should receive different noise levels. Edited regions need intense noise to disrupt both frequency components for effective conversion to target content, while unedited regions should use low noise to retain original information. Inappropriate use of inversion noise limits the quality of video editing with these tuning-free methods. 




In this paper, we propose a tuning-free instruction-based video editing model to complete efficient subject or attribute replacement and deletion with refined noise control. Specifically, we design the Edit Instruction Analysis Module (EIAM) to describe the source video with the source prompt $\mathcal{P}_{\mathcal{S}}$, reason edited video with the target prompt $\mathcal{P}_{\mathcal{T}}$ through Chain-of-thought (CoT), and locate the edited area with mask $\mathcal{M}$. We propose a Structural Noise Initialization Strategy (SNIS): the edited area starts with a higher noise level to change the content in source video while the unedited area starts with a lower noise level to retain the content. Structured and differentiated initial noise helps fulfill the editing instruction and preserve the unedited content. To ensure the coordination of the edited video and preservation of unedited content, we introduce a systematic Noise Guidance Mechanism (NGM) in the denoising process in video diffusion models. 


Our contributions can be summarized as follows: 
\begin{itemize}

\item (1) We design an instruction-based video editing model without any fine-tuning, which supports both replacement and removal in the object-level and attribute-level. 
\item (2) We propose a Structural Noise Initialization Strategy (SNIS) and a Noise Guidance Mechanism (NGM) for unedited areas, extending generation capabilities to editing capabilities from a noise perspective. 
\item (3) Experimental results show our model achieves better visual quality and state-of-the-art performance.
\end{itemize}

\begin{figure*}[t]
    \begin{center}
    \includegraphics[width=\linewidth]{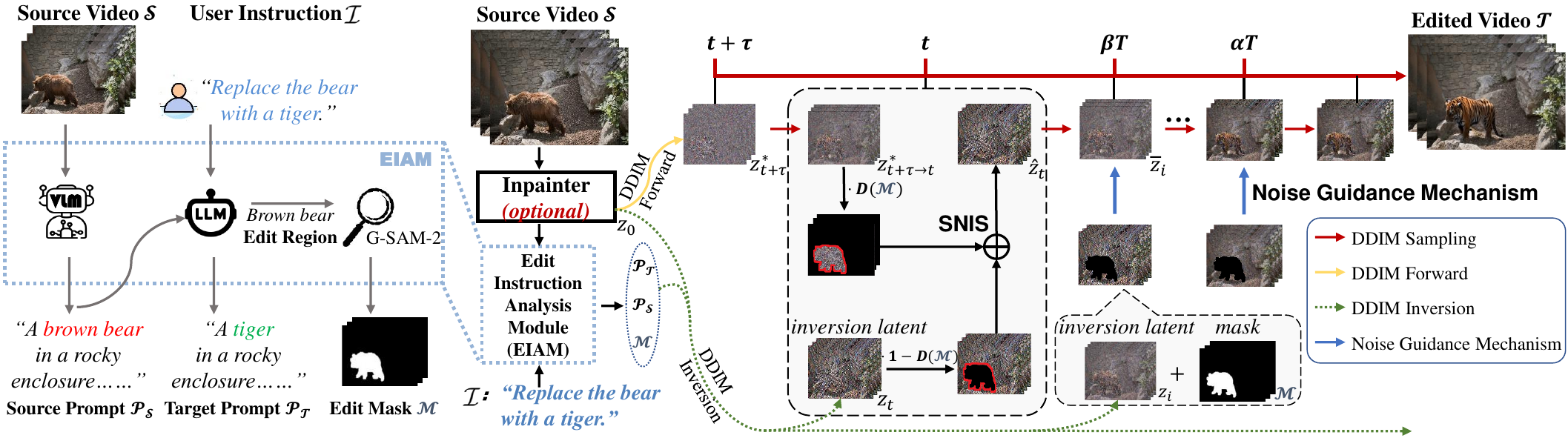}
    \end{center}
    \vspace{-0.3cm}
     \caption{The overview of our proposed methods. Utilizing the tailored framework, which incorporates EIAM, SNIS and NGM, source video $\mathcal{S}$ can be transformed into target video $\mathcal{T}$ according to user instruction $\mathcal{I}$. G-SAM-2 denotes Grounded-SAM-2. At timestep $t$, we introduce the SNIS to get a  denoised starting state $\hat{z}_{t}$. During the period $i\in[\alpha T, \beta T]$, we introduce the NGM to maintain the background and optimize the generated video by leveraging video priors.}
     \label{fig:pipeline}
    \vspace{-0.2cm}
 \end{figure*}

\section{Related Works} 
\label{sec:peerworks}

Relevant works in image editing focus on converting image generation models into editing models through prompt guidance and attention manipulation~\cite{InstructPix2Pix, P2P, Plug-and-Play}. Owing to the delayed development of video generation models~\cite{HunyuanVideo,CogVideoX,WanX} relative to image generation models~\cite{SD}, early video editing research focused on customizing image editing techniques for video editing tasks. This customization is typically achieved by incorporating spatio-temporal attention mechanisms~\cite{Text2Video-Zero, FateZero, Pix2Video, Tune-A-Video} or integrating additional motion information into image generation models~\cite{TokenFlow, VideoShop, DIVE, VideoSwap}. While these tailored techniques alleviate the temporal information deficit in image generation models, edited videos still exhibit temporal inconsistency. This issue remains a critical bottleneck for achieving high-quality video editing performance.

Subsequent efforts attempted to convert video generation backbones~\cite{HunyuanVideo,CogVideoX,WanX} with enhanced temporal consistency into video editing models, which fall into two technical routes. Among them, training-based methods~\cite{VACE, MiniMax} leverage carefully curated datasets to fine-tune video generation models~\cite{HunyuanVideo, CogVideoX, WanX}. However, their practical applicability relies on large-scale high-quality datasets and intricate fine-tuning pipelines. In contrast, tuning-free methods~\cite{AnyV2V, V2Edit} seek to customize inversion mechanisms and attention manipulation techniques, thereby circumventing the high costs associated with model fine-tuning. Importantly, high- and low-frequency information exhibit distinct corruption rates during the diffusion process~\cite{FreeInit}. This property necessitates differentiated noise levels for edited and unedited regions: edited regions require intense noise to disrupt both frequency components, facilitating effective conversion to the target content, while low noise should be applied to unedited regions to preserve the original information. Inappropriate use of inversion noise limits the quality of video editing with these tuning-free methods.

\section{Methods} 
\label{sec:methods}

This paper proposes an instruction-driven video editing framework, which supports object or attribute replacement and deletion. We will discuss the proposed Edit Instruction Analysis Module (EIAM), Structural Noise Initialization Strategy (SNIS) and Noise Guidance Mechanism (NGM).


\subsection{Edit Instruction Analysis Module}
\label{subsec_2_2}
This paper constructs a video editing model based on a video generation model, adapting it for precise editing to fulfill user instructions and preserve unedited content. To reduce reliance on user-provided detailed descriptions, we propose an Edit Instruction Analysis Module (EIAM), which analyzes the instruction $\mathcal{I}$ and source video $\mathcal{S}$. 


Specifically, inspired by VLMs in automatic labeling of visual data, we employ a VLM to process the source video and obtain its corresponding prompt $\mathcal{P}_{\mathcal{S}}$. After obtaining $\mathcal{P}_{\mathcal{S}}$, we combine it with user instruction $\mathcal{I}$ to infer the expected state of the edited video based on the LLM with enhanced logical reasoning abilities via Chain-of-Thought (CoT). Then, we get the target prompt $\mathcal{P}_{\mathcal{T}}$ and the name of the edit object(s). Our model supports both object-level and attribute-level edits. Assuming the user specifies the edit region as the area where one or multiple objects appear in the video, the task becomes a classic object tracking problem. We use Grounded-SAM-2~\cite{GSAM2} to segment the target region based on the identified object name, producing a precise mask $\mathcal{M}$ that distinguishes between edited and unedited areas. As shown in Fig.~\ref{fig:pipeline}, EIAM outputs a source prompt $\mathcal{P}_{\mathcal{S}}$, a target prompt $\mathcal{P}_{\mathcal{T}}$, and an edit mask $\mathcal{M}$ indicating the edit regions.

\subsection{Structural Noise Initialization Strategy (SNIS)}
\label{subsec_2_3}
In tuning-free video editing, pre-trained video generation models are repurposed to achieve controllable editing. Given  $\mathcal{P}_{\mathcal{T}}$ from the proposed EIAM, the core challenge lies in guiding the generative model to execute the edit, preserve unedited content, and maintain spatiotemporal consistency in the output video $\mathcal{T}$. Existing approaches typically invert the source video $\mathcal{S}$ into a noisy latent via DDIM inversion~\cite{DDIM} and control denoising through attention manipulation. However, these methods suffer from two limitations: (1) applying a uniform noise level across the entire latent ignores regional distinctions—edited regions require stronger noise to facilitate change, while unedited areas need weaker noise to retain details; (2) the content of unedited regions is encoded in the noisy latent rather than in attention maps or queries.

As shown in Fig.~\ref{fig:pipeline}, we add noise to the clean latent $z_0$ ($x_0$) in two ways: using the DDIM forward process and the DDIM inversion process. We get two noisy latents: $z_{t+\tau}^{*}$ with random noise at timestep $t+\tau$ ($\tau \ge 0$), and $z_t$ with inversion noise at timestep $t$. We use $z_{t+\tau}^{*}$ to initialize the edited regions, and $z_t$ for the unedited regions. There are two reasons for this setting: (1) Using random noise aligns with how generative models are trained, where source videos are perturbed with random noise. This helps ensure that content generated in the edited regions closely follows the user’s instructions through the video generation model. (2) inversion noise helps preserve structural and detail consistency in unedited regions. We also keep the noise level in $z_{t+\tau}^{*}$ higher than in $z_t$. Higher noise makes it easier to change content in edited regions, while lower noise helps preserve unedited areas. However, simply combining these two noisy latents leads to artifacts and mosaic effects in the edited video. Thus, we first denoise $z_{t+\tau}^{*}$ to an intermediate state $z_{t}^{*}$ with the same noise level as $z_t$ according to the Eq.~\eqref{eq_reverse}. The equation 
\begin{equation}
    \overline{z}_{t-1} = \overline{z}_{t} - \epsilon_{\theta}(\overline{z}_{t}, \mathcal{P}_{\mathcal{T}}) \label{eq_reverse}
\end{equation}
denotes the generation model completes one denoising process. Inspired by the latent update of Pix2Video~\cite{Pix2Video}, we introduce a transition zone {(marked red in Fig.~\ref{fig:pipeline})} to facilitate the fusion of the two types of noisy latent. Within the transition zone, a linear decay strategy is applied: the closer to the edited region, the larger the weighting coefficient, and the greater the proportion of $z_{t}^{*}$ included. Our proposed SNIS at timestep $t$ can be formulated as:
\begin{equation}
    \hat{z_t} = D\left(\mathcal{M}\right)\cdot z_{t}^{*} + \left(1 - D\left(\mathcal{M}\right) \right)\cdot z_t \label{eq_init}.
\end{equation}
The $D(\cdot)$ is a coefficient distribution matrix associated with the mask $\mathcal{M}$. The $D(\cdot)$ is defined as:
\begin{equation}
    D(\mathcal{M})(x, y) = \left\{
    \begin{array}{cc}
        \frac{\text{max}(m-d, 0)}{m},  & d \in [0, m]  \\
        1 , & otherwise 
    \end{array} ,
    \right. \label{eq_d()}
\end{equation}
where $d$ denotes the minimal distance from the point $(x, y)$ to the edited regions indicated by the mask $\mathcal{M}$, and $m$ is a hyperparameter that controls the width of the transition zone.
 
Unless initialized as Gaussian noise, $z_{t+\tau}^{*}$ may retain edited objects' information, causing artifacts. We find that while using pure Gaussian noise reduces such retention, it can introduce distortions near editing boundaries. 
To improve visual quality, we \textbf{optionally} introduce an inpainting model for preprocessing~\cite{SuvorovLMRASKGP22}. This is particularly applied in removal and replacement tasks with structural discrepancies. Specifically, it alleviates structural biases in replacement tasks, and effectively mitigates residual information in removal tasks.


\subsection{Noise Guidance Mechanism (NGM)}
\label{subsec_2_4}

Our SNIS (Eq.~\eqref{eq_init}) produces a denoising state $\hat{z_t}$ with region-adaptive noise levels for edited and unedited areas. Inspired by RAGD~\cite{RAG}, where noisy latent per object are combined using global prompts to improve coherence, one may directly use $\hat{z_t}$ for denoising guided by $\mathcal{P}_{\mathcal{T}}$. However, we observe that at large timesteps, significant background discrepancies occur even in reconstructions from $z_t$ alone, complicating the preservation of background consistency during editing.

To address this issue, we propose a Noise Guidance Mechanism (NGM) that uses intermediate noisy latent $\mathcal{Z}=\left\{z_i | i\in[\alpha T, \beta T] \right\}$ with lower noise levels to gradually steer the denoising network toward generating consistent and harmonious video edits. The NGM after each denoising step shown in Eq.~\eqref{eq_reverse} can be formulated as:
\begin{equation}
    \overline{z}_{i} = \left\{
    \begin{array}{cc}
        \mathcal{M} \cdot \overline{z}_{i} + (1 - \mathcal{M}) \cdot z_i,  & i \in [\alpha T, \beta T]  \\
        \overline{z}_{i} , & otherwise 
    \end{array} .
    \right. \label{eq_NGM}
\end{equation}
The noisy latent $z_i$ in $\mathcal{Z}$ is the noisy latent obtained by $z_0=x_0$ through $i$ inversion processes. The symbol $T$ denotes the total inference steps for the video generation models. The factors $\alpha$ and $\beta$ control the range of NGM. 

\begin{figure*}[h]
    \begin{center}
    \includegraphics[width=\linewidth]{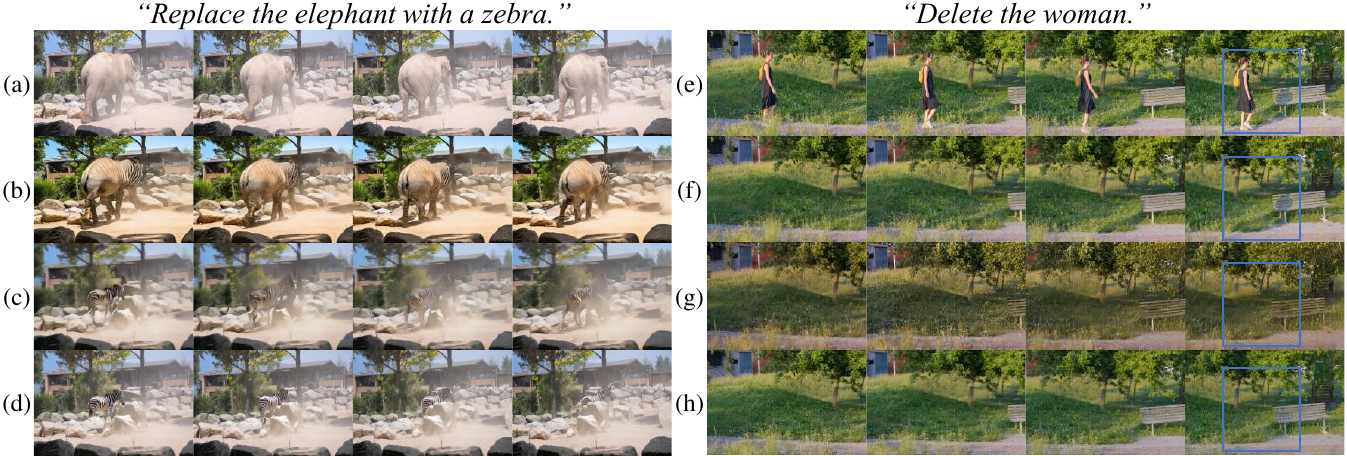}
    \end{center}
    \vspace{-0.1cm}
     \caption{Qualitative comparison with peer methods. The video (a) and (e) denote source video while the other video are edited video from (b) TokenFlow, (f) MiniMax-Remover, (c)\&(g) AnyV2V, (d)\&(h) our model. }
     \label{fig:peer_methods}
    \vspace{-0.6cm}
 \end{figure*}

We do not use our proposed NGM throughout the entire process. {The reason is that predicted masks from models like Grounded-SAM-2 tend to segment only the object, failing to capture environmental effects such as shadows or reflections.} By allowing joint denoising in a range from the timestep $\alpha T$ to the timestep 0 (where inversion error is small), the model leverages its inherent video priors to refine the environmental effects like shadows and reflections, even outside the explicitly masked region, thereby enhancing overall consistency. 

\section{Experiments} 
\label{sec:experiments}

\subsection{Experimental Setup}
\label{subsec:setup}

We employ CogVideoX-5B~\cite{CogVideoX} as the video generation model in this paper. In the proposed EIAM, InternVL2.5-26B~\cite{internVL25} is utilized as the VLM to obtain source prompt while Qwen3-72B~\cite{Qwen3} serves as the LLM to generate target prompt and identify the target subject for editing. Furthermore, Grounded-SAM-2~\cite{GSAM2} is applied to acquire masks of the edited regions. Regarding hyperparameter configuration, we empirically set $\tau = 5$, the total inference steps $T$ to $100$, and the width of transition zone $m=16$. The coefficient $\alpha$ and $\beta$ is set to $0.48$ and $0.85$, respectively, so as to effectively preserve unedited regions and enhance the overall efficiency of the editing process.

\vspace{-0.5cm}

\subsection{Experimental Datasets and Metrics}
\textbf{Datasets}: 
To evaluate our approach in real-world scenarios, like previous work \cite{TokenFlow, Pix2Video}, we constructed a dataset based on the DAVIS~\cite{DAVIS}, which consists of high-quality, real-video sequences. Each instance includes an original video and a corresponding edit instruction. The tasks are categorized into: object removal, object replacement, and object attribute modification. Specifically, the dataset covers a diverse range of challenges, including structurally similar replacement (encompassing attribute modification) and structurally dissimilar replacement. Note that our model does not require any training data.

\setlength{\parindent}{0pt}{\textbf{Metrics}: We quantitatively evaluate our model and other works from three aspects: (i) \textit{Instruction-following capability}. Considering the difficulty in directly assessing the consistency between user instructions and the edited video, we evaluate this capability by measuring the semantic alignment between the target prompt and the edited video via the CLIP-T~\cite{AnyV2V}. (ii) \textit{Unedited regions preservation capability}. To avoid inaccuracies in metric evaluation caused by artifacts such as shadows and reflections after shadow removal, we drop pixel-level metrics like PSNR and SSIM, and instead adopt LPIPS~\cite{LPIPS} to evaluate perceptual similarity. 
{(iii)Video generation quality. Following common practices in video generation evaluation, we adopt Fréchet Video Distance (FVD)~\cite{2018Towards} to measure the distribution-level similarity between generated videos and real videos in the spatiotemporal feature space. In addition, we use CLIP-I~\cite{AnyV2V} to assess the temporal consistency of the generated videos, as widely adopted in prior works~\cite{AnyV2V,Pix2Video}.}


\vspace{-0.5cm}
\subsection{Experimental Results}

We compare our method with several state-of-the-art baselines: \red{Pix2Video~\cite{Pix2Video}, TokenFlow~\cite{TokenFlow} and AnyV2V~\cite{AnyV2V}} support object and attribute replacement; \red{AnyV2V and MiniMax-Remover~\cite{MiniMax}} support object removal. We evaluate these methods on both tasks. As shown in \red{Tab.~\ref{tab_compared}}, our method outperforms all previous approaches. 

Our performance gains are attributed to three key factors: 
First, our SNIS ensures precise instruction adherence by selectively suppressing noise in unedited areas to maintain structure while amplifying it in edited areas to induce change. 
Second, with NGM, our method leverages rich information in noisy latents to steer the denoising process and preserve unedited content. 
Third, by leveraging the generation model's inherent video priors, we jointly denoise both regions without explicit noise guidance during the later steps.

Qualitative comparisons in \red{Fig.~\ref{fig:peer_methods}} validate these advantages. Our method effectively alleviates biases caused by structural discrepancies (e.g., between \red{the elephant and zebra}) and mitigates residual information (e.g., of \red{the woman}). In contrast, prior methods like TokenFlow and AnyV2V often struggle to break the structural priors of the source video in replacement tasks, resulting in low-quality generation when significant shape discrepancies exist. Furthermore, in removal tasks, AnyV2V exhibits poor background consistency, while MiniMax-Remover fails to eliminate environmental effects, often leaving behind visible artifacts such as cast shadows.






\begin{table}[t]
\centering
\caption{Quantitative comparison with previous methods.} 
\begin{tabular*}{\linewidth}{@{\extracolsep{\fill}}lcccc}
\hline
\hline
Method         & CLIP-T$\uparrow$ & LPIPS$\downarrow$ & FVD$\downarrow$ & CLIP-I$\uparrow$  \\
\hline
\multicolumn{5}{c}{Object and attribute replacement} \\
\hline
Pix2Video~\cite{Pix2Video}    & {0.3174} & 0.5621 & 1547.33  & 0.9774  \\
TokenFlow~\cite{TokenFlow}   & {0.3196} & 0.3471 & 1049.94  & \underline{0.9803}    \\     
AnyV2V~\cite{AnyV2V}   & \underline{0.3203} & \underline{0.3238} & \underline{740.99}  & 0.9641    \\     
Ours    & \textbf{0.3249} & \textbf{0.1526} & \textbf{449.02}  & \textbf{0.9823}      \\   
\hline
\multicolumn{5}{c}{Object removal} \\
\hline
AnyV2V~\cite{AnyV2V}   & \underline{0.2994} & 0.3643 & 926.34  & 0.9655    \\     
M-Remover~\cite{MiniMax} & 0.2961 & \textbf{0.1446} & \textbf{518.18}  & \underline{0.9761}  \\ 
Ours    & \textbf{0.3048} & \underline{0.1828} & \underline{618.00}  & \textbf{0.9826}      \\   
\hline
\hline
\end{tabular*}\label{tab_compared}
   \raggedright{
   \begin{tablenotes}
      \footnotesize
      \item[*] 1. Best and second scores are \textbf{highlighted} and \underline{underlined} respectively.
    \end{tablenotes}
    }
    \vspace{-0.3cm}
\end{table} 


 \subsection{Discussion and Analysis}
\label{subsec:analysis}

\textbf{Robustness against Mask Inaccuracies.} 
In video editing pipelines involving segmentation, errors in masks (e.g., from Grounded-SAM-2) typically propagate into the editing process, leading to failures or visual artifacts. A common issue is that shadows and reflections of the target object often fall outside the generated mask (described in Sec.~\ref{subsec_2_4}).
However, our proposed NGM mitigates this issue by leveraging inherent video priors during the partial denoising intervals to correct content even outside the explicit mask regions. 
In comparison, baseline methods like MiniMax-Remover strictly limit changes to the masked area (See Fig.~\ref{fig:peer_methods} right).Consequently, they fail to remove residual shadows or reflections in non-edited regions in almost all removal cases. 
This analysis further explains why our method may score lower on pixel-level metrics like LPIPS (which penalize background changes) but is preferred by human evaluators for its visual plausibility and cleanliness.

\begin{table}[t]
\centering
\caption{Ablation Studies of proposed methods.} 
\begin{tabular*}{\linewidth}{@{\extracolsep{\fill}}lcccc}
\hline
\hline
Method         & CLIP-T$\uparrow$ & LPIPS$\downarrow$ & FVD$\downarrow$ & CLIP-I$\uparrow$  \\
\hline
Ours    & \underline{0.3153} & \textbf{0.1669} & \textbf{370.88}  & \underline{0.9824}  \\
$w/o$ NGM   & \textbf{0.3240} & 0.5139 & 621.31  & \textbf{0.9879}    \\     
$w/o$ SNIS   & 0.3126 & \underline{0.1901} & \underline{463.95}  & 0.9805    \\     
\hline
\hline
\end{tabular*}\label{tab_ablation}
\end{table} 


\subsection{Ablation Studies}

To comprehensively evaluate the effectiveness and necessity of each component within our proposed framework, we conduct a series of quantitative and qualitative ablation studies, with results summarized in \red{Tab.~\ref{tab_ablation}}. 

\textbf{Impact of Noise Guidance Module (NGM).} 
First, we investigate the contribution of the NGM, which is designed to regulate the inversion noise during the critical denoising interval $[\alpha T, \beta T]$. As illustrated in the qualitative results of \red{Fig.~\ref{fig:ablation}} ($w/o$ NGM), removing the NGM leads to a noticeable degradation in the preservation of unedited regions. Although the semantic alignment (CLIP-T) remains relatively high, this comes at the cost of structural fidelity. Quantitative metrics in \red{Tab.~\ref{tab_ablation}} corroborate this observation: the absence of NGM results in a significant increase in LPIPS scores (indicating lower similarity to the source) and FVD. This suggests that NGM acts as a vital structural anchor, steering the generation process to respect the original video layout while preventing the model from hallucinating excessive changes in the background.

\textbf{Impact of SNIS Module.} 
Next, we assess the role of the Selective Noise Injection/Suppression (SNIS) mechanism. As shown in \red{Tab.~\ref{tab_ablation}}, removing SNIS leads to a sharp decline in the CLIP-T score, which reflects a weakened instruction-following capability. Visual inspection in \red{Fig.~\ref{fig:ablation}} ($w/o$ SNIS) further reveals that without the targeted noise amplification provided by SNIS, the model struggles to fully overwrite the original object features, resulting in "under-editing" issues or residual artifacts (e.g., ghosting of the source object). These results underscore that SNIS is indispensable for effectively disrupting the source signal in the edit region to facilitate precise semantic modification.



\begin{figure}[t]
    \begin{center}
    \includegraphics[width=\linewidth]{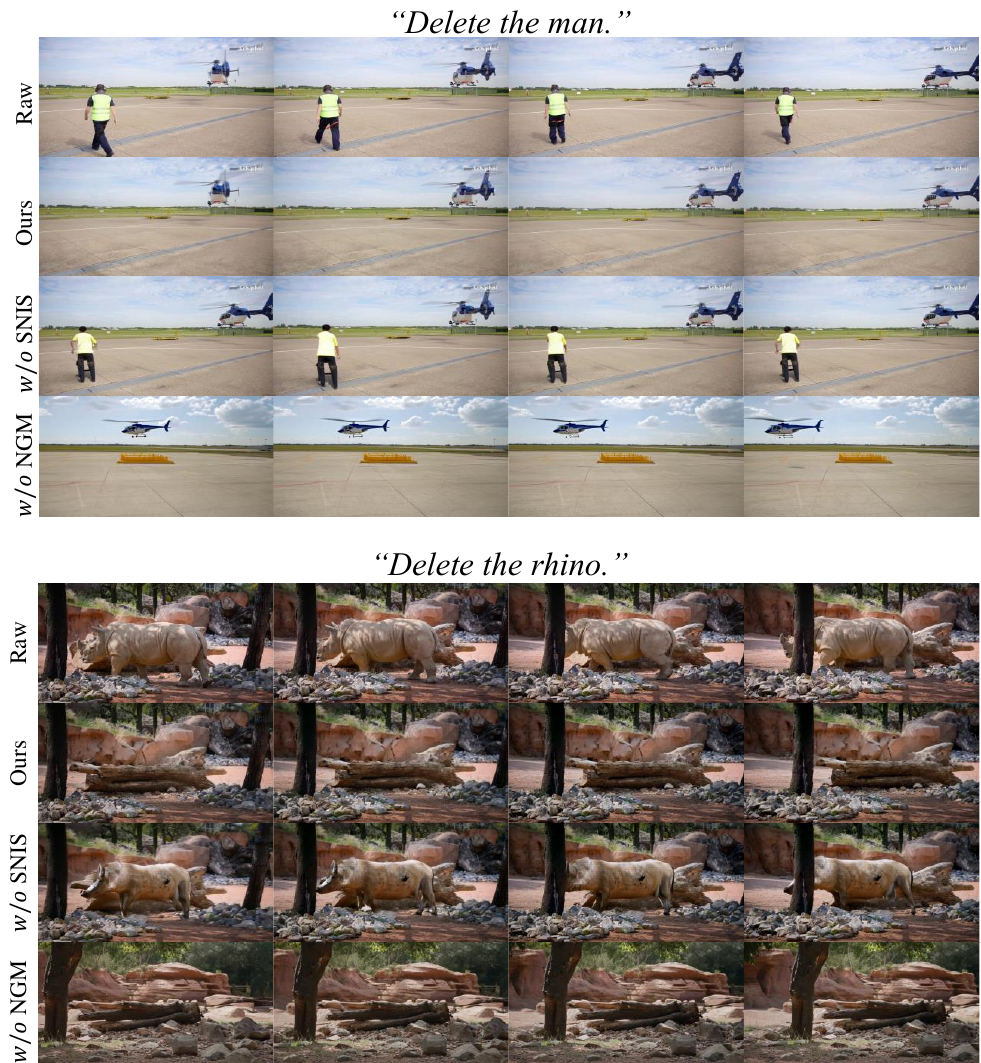}
    \end{center}
     \vspace{-0.6cm}
     \caption{Qualitative results of ablation studies.}
     \label{fig:ablation}
    \vspace{-0.2cm}
 \end{figure}

\vspace{-0.5cm}

\section{Conclusion}
\label{sec:conclusion}

In this paper, we propose a tuning-free and instruction-driven video editing framework. Specifically, the EIAM is used to analyze the edit instruction and input video.  We propose the SNIS that initializes the diffusion denoising process with spatially varying noise levels. Furthermore, the NGM is introduced to leverage rich information in noisy latent and the inherent video priors in generative model to guide denoising while maintaining the coherence of unedited content and overall visual integrity. 

\bibliographystyle{IEEEbib}
\bibliography{strings,refs}

\end{document}